\documentclass[preprint,12pt]{elsarticle}

\usepackage{amssymb}
\usepackage{amsmath}
\usepackage{graphicx}
\usepackage{booktabs}
\usepackage{adjustbox}
\usepackage{float}

\journal{}

\begin{document}

\begin{frontmatter}

\title{Multimodal Posterior Sampling-based Uncertainty in PD-L1 Segmentation from H\&E Images}

\tnotetext[t2]{\textit{Preprint.} This manuscript has been accepted for publication in 
\textbf{Lecture Notes in Bioinformatics (Springer, 2025)}. }

\author[uc3m]{Roman Kinakh\corref{cor1}}
\ead{rkinakh@ing.uc3m.es}

\author[uc3m,iisgm]{Gonzalo R. Rios-Mu\~noz}
\ead{grios@ing.uc3m.es}

\author[uc3m]{Arrate Mu\~noz-Barrutia}
\ead{mamunozb@ing.uc3m.es}

\cortext[cor1]{Corresponding author.}

\affiliation[uc3m]{
  organization={Universidad Carlos III de Madrid},
  addressline={Av. de la Universidad, 30},
  city={Legan\'es (Madrid)},
  postcode={28911},
  country={Spain}
}

\affiliation[iisgm]{
  organization={Instituto de Investigaci\'on Sanitaria Gregorio Mara\~n\'on},
  city={Madrid},
  country={Spain}
}

\begin{abstract}
Accurate assessment of PD-L1 expression is critical for guiding immunotherapy, yet current immunohistochemistry (IHC) based methods are resource-intensive. We present nnUNet-B: a Bayesian segmentation framework that infers PD-L1 expression directly from H\&E-stained histology images using Multimodal Posterior Sampling (MPS). Built upon nnUNet-v2, our method samples diverse model checkpoints during cyclic training to approximate the posterior, enabling both accurate segmentation and epistemic uncertainty estimation via entropy and standard deviation. Evaluated on a dataset of lung squamous cell carcinoma, our approach achieves competitive performance against established baselines with mean Dice Score and mean IoU of 0.805 and 0.709, respectively, while providing pixel-wise uncertainty maps. Uncertainty estimates show strong correlation with segmentation error, though calibration remains imperfect. These results suggest that uncertainty-aware H\&E-based PD-L1 prediction is a promising step toward scalable, interpretable biomarker assessment in clinical workflows.
\end{abstract}




\end{frontmatter}
\section{Introduction}

Programmed death-ligand 1 (PD-L1) is a transmembrane protein expressed on tumor and immune cells that plays a key role in suppressing the immune response~\cite{topalian2012safety}. Its expression is a critical biomarker for identifying patients likely to benefit from immune checkpoint inhibitors, a class of cancer immunotherapies~\cite{pardoll2012blockade,ribas2015releasing}. The accurate assessment of PD-L1 expression in tumor tissue is essential for guiding immunotherapy decisions across various cancers~\cite{ai2020roles}. Traditionally, PD-L1 is evaluated using immunohistochemistry (IHC), which directly visualizes protein expression. While clinically effective, IHC is resource-intensive, time-consuming, and subject to inter-observer variability. In contrast, Hematoxylin and Eosin (H\&E) staining is a standard, inexpensive, and widely available diagnostic modality. This work explores the feasibility of segmenting PD-L1-expressing tumor regions directly from H\&E-stained images, potentially offering faster, more accessible alternatives for patient stratification~\cite{wang2024prediction}.

Histology, as a medical imaging domain, presents a unique set of challenges that render both image analysis and clinical decision-making difficult. As illustrated in Fig.~\ref{fig:dataset}, the microscopic environment of tissue sections is inherently complex and heterogeneous~\cite{fuchs2011computational}. Unlike the often distinct and macroscopic structures seen in tomographic scans, cells and tissue types in H\&E and IHC images are densely packed, exhibit highly diverse morphologies at a microscopic level, and can be organized in intricate, overlapping, and often ambiguous patterns~\cite{marini2021deep}. Furthermore, subtle variations in tissue processing, staining protocols, and imaging conditions introduce substantial inter-slide variability that can significantly impact model robustness~\cite{tellez2021histo,stacke2021measuring}. These factors contribute to the "noisy" and ambiguous nature of histology data, demanding advanced computational methods that can discern subtle yet critical biological signals amidst a sea of microscopic complexity, such as epistemic uncertainty estimation.


While techniques like Monte Carlo Dropout (MCDO)~\cite{gal2016dropout} have been widely used for uncertainty estimation in medical imaging, they often lead to a trade-off between prediction confidence and segmentation accuracy. This trade-off is particularly problematic in histology, where fine-grained errors can impact downstream biomarker quantification. To address this, we introduce nnUNet-B: a Multimodal Posterior Sampling (MPS) framework based on nnUNet-v2 ~\cite{isensee2021nnu,isensee2024nnu} that provides richer, more stable uncertainty estimates without degrading segmentation performance. Our approach better captures model variability while maintaining a computational cost comparable to MCDO, offering clinicians a clearer picture of where and why the model may be uncertain — an essential feature for deploying AI in sensitive diagnostic workflows.

\section{Materials and Methods}

\subsection{Dataset}\label{subsec:dataset}

\begin{figure}[ht!]
    \centering
    \includegraphics[width=\textwidth]{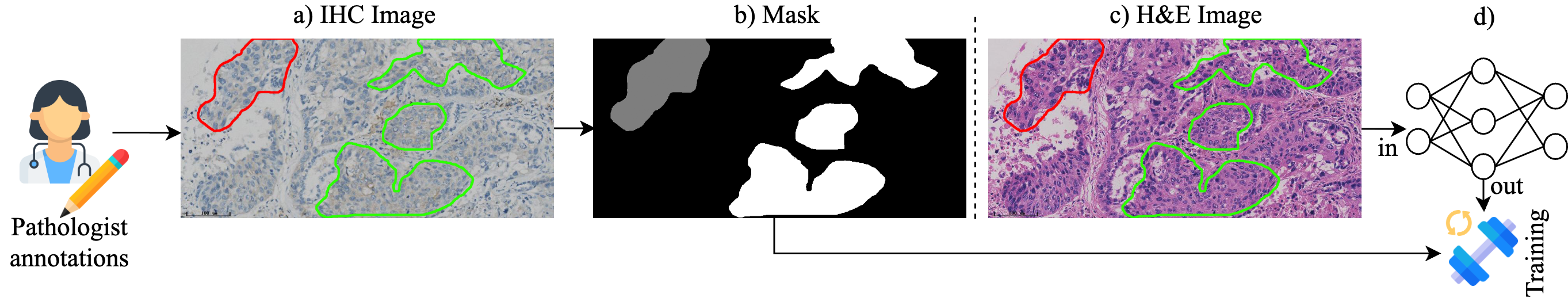}
    \caption{Overview of dataset annotation and segmentation workflow: (a) Pathologists annotate PD-L1-positive (green) and PD-L1-negative (red) tumor regions on IHC slides. (b) Annotations are converted into 3-class segmentation masks. (c) Masks are aligned with corresponding H\&E images. (d) H\&E images are used as model input, with predictions supervised by the aligned PD-L1 masks. \cite{deng2025mcranet,wang2024prediction}}
    \label{fig:dataset}
\end{figure}

To train and evaluate our model, we used the dataset introduced by Wang et al.~\cite{wang2024prediction}, which comprises 1,088 paired H\&E- and IHC-stained histology images of lung squamous cell carcinoma. Each H\&E image is annotated with pixel-wise PD-L1-positive and -negative tumor regions, using the corresponding IHC slide as reference. The annotation process is illustrated in Fig.~\ref{fig:dataset}.

For our experiments, we randomly allocated 20\% of the images (218) as a held-out test set. From the remaining 80\% (870 images), we used 20\% (174 images) for validation and the remaining 696 images for training. All images have a size of $959 \times 923$ pixels with the pixel size of approximately 1.5 $\mu m$.

\subsection{Model Architecture}

\begin{figure}[ht!]
    \centering
    \includegraphics[width=\textwidth]{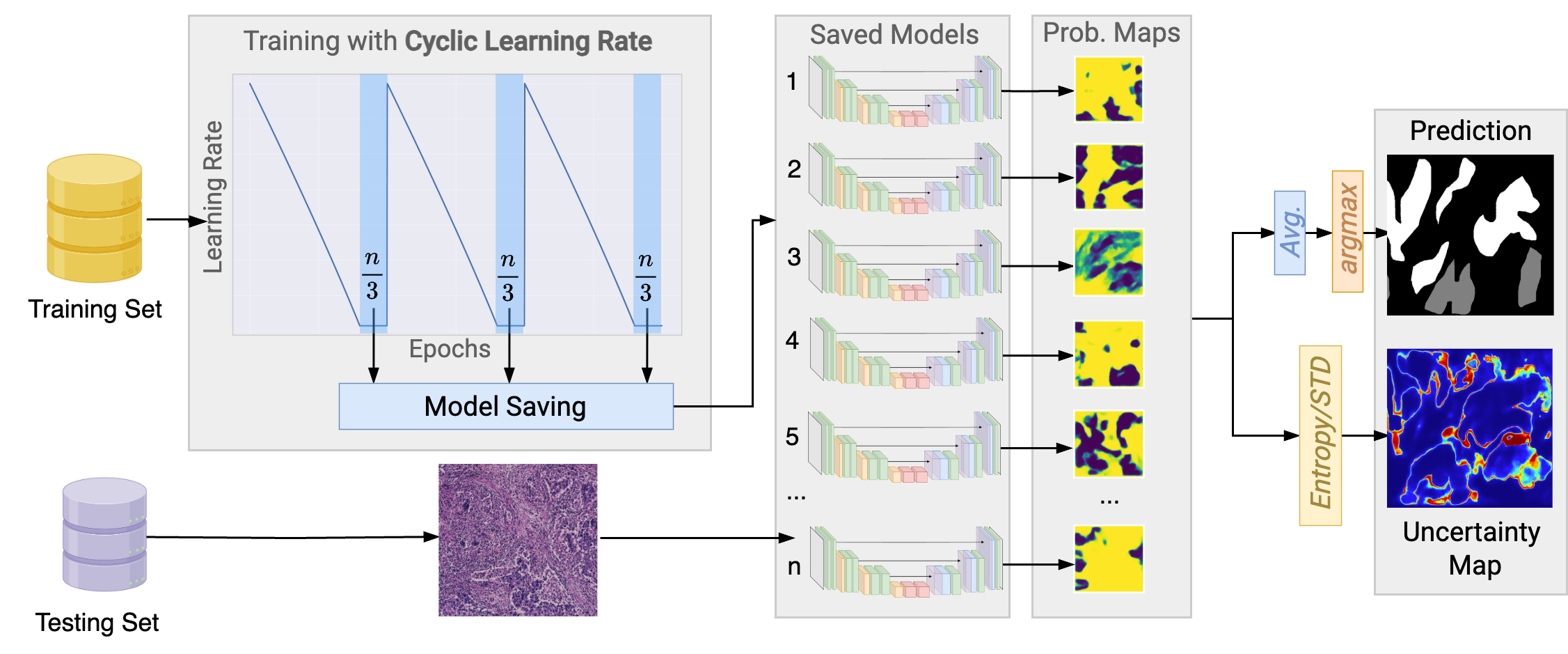}
    \caption{Bayesian nnU-Net framework with Multimodal Posterior Sampling (MPS). During training, checkpoints are sampled from the last $n/3$ epochs of each learning cycle. At inference, an H\&E image is passed through $n$ sampled models to generate probability maps, which are averaged and $\arg\max$-ed for prediction. Pixel-wise uncertainty is computed using entropy or standard deviation. \cite{zhao2022efficient}}
    \label{fig:architecture}
\end{figure}

The nnUNet-B method is based on the nnUNet-v2 architecture~\cite{isensee2021nnu}, which serves as the segmentation backbone. To enable uncertainty-aware predictions, we extended it using the MPS strategy introduced by Zhao et al.~\cite{zhao2022efficient}. In this approach, multiple model instances sampled from different local minima of the optimization trajectory are treated as approximate posterior samples (Fig.~\ref{fig:architecture}). These samples are obtained from saved checkpoints during different phases of training (detailed in Section~\ref{subsec:training}).

At inference time, each sampled model $\mathcal{M}_i$ produces a softmax probability map $P_i(\mathbf{x}) \in [0, 1]^C$ for an input image $\mathbf{x}$, where $C$ is the number of segmentation classes. The ensemble of $N$ such models yields a set $\{P_i(\mathbf{x})\}_{i=1}^{N}$. The final prediction is computed by averaging the probabilities across all models: $\bar{P}(\mathbf{x}) = \frac{1}{N} \sum_{i=1}^{N} P_i(\mathbf{x})$.

The predicted segmentation mask $\hat{y}$ is obtained by taking the voxel-wise arguments of the maxima over the averaged probabilities: $\hat{y} = \arg\max_c \bar{P}_c(\mathbf{x})$. To quantify predictive uncertainty, we computed two pixel-wise measures over the ensemble: the standard deviation (STD, $\sigma$) and the entropy ($H$) of the averaged distribution:

\begin{equation}
\begin{array}{ll}
\sigma(\mathbf{x}) = \sqrt{ \frac{1}{N} \sum_{i=1}^{N} (P_i(\mathbf{x}) - \bar{P}(\mathbf{x}))^2 }, & 
H(\mathbf{x}) = - \sum_{c=1}^{C} \bar{P}_c(\mathbf{x}) \log \bar{P}_c(\mathbf{x})
\end{array}
\end{equation}
This design enables the extraction of both the most probable segmentation and its associated uncertainty, without modifying the underlying network architecture or requiring stochastic components at inference time.

\subsection{Model Training}\label{subsec:training}

\begin{figure}[ht!]
    \centering
    \includegraphics[width=\textwidth]{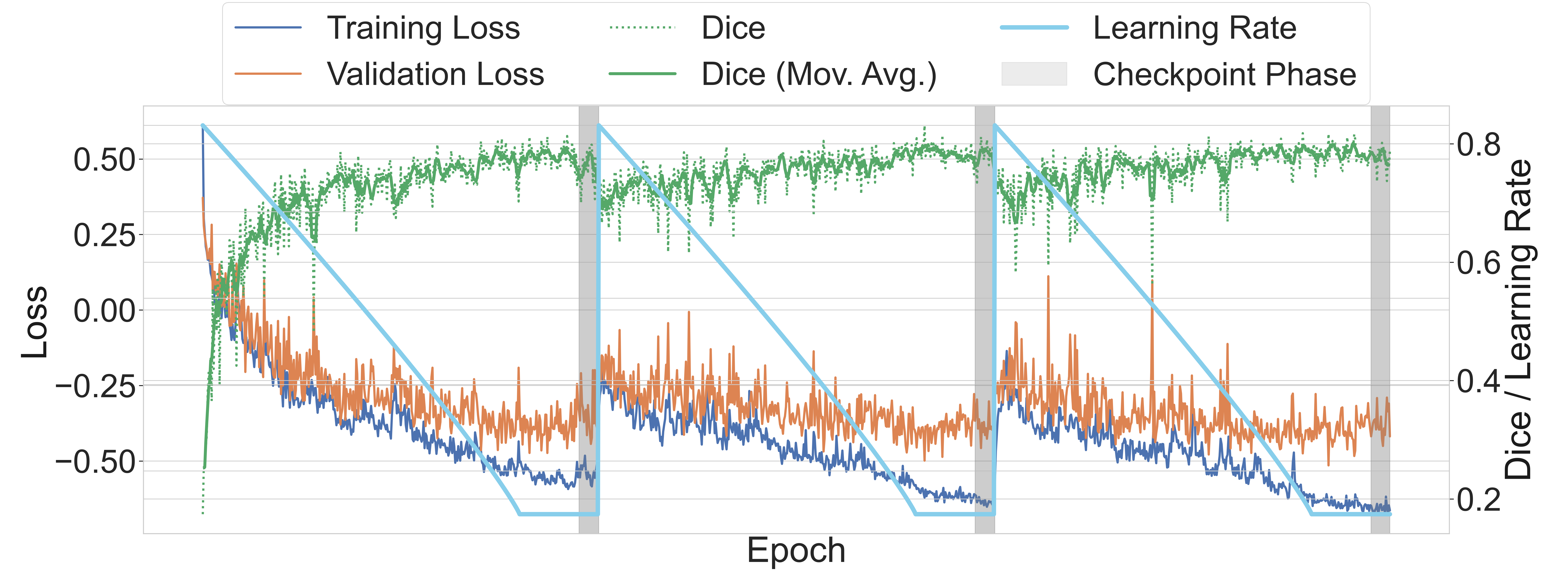}
    \caption{Training process of the Bayesian nnU-Net framework with Multimodal Posterior Sampling (MPS).
The model undergoes three full training cycles using a cyclic learning rate schedule. During the final 20 epochs of each cycle, model checkpoints are sampled and stored to later be used as an ensemble for uncertainty estimation.}
    \label{fig:training}
\end{figure}

The model was trained similarly to a standard U-Net using the combination of \textbf{Dice} and \textbf{Cross-Entropy} losses. To promote convergence while preserving exploratory behavior across training, we employed a \textbf{cyclical learning rate (CLR)} schedule based on polynomial decay \cite{smith2017cyclical}. At the beginning of each cycle of length $T_c$, the learning rate is initialized to a higher value \( \alpha_r \), and then decays polynomially to a minimum value \( \alpha_0\) over a fraction \( \gamma \) of the cycle length, with a polynomial decay power of \( \epsilon\). Beyond this point, the learning rate remains constant at \( \alpha_0 \) for the remainder of the cycle, which ensures stability during later iterations while enabling aggressive updates early in the cycle.  Therefore, we define the learning rate \( \alpha(t) \) for epoch \( t \) as: 

\begin{equation}
\label{eq:clr}
\alpha(t) =
\begin{cases}
\alpha_0 + (\alpha_r - \alpha_0) \left(1 - \dfrac{t_c}{\gamma T_c} \right)^{\epsilon}, & \text{if } 0 < t_c \leq \gamma T_c \\
\alpha_0, & \text{if } t_c > \gamma T_c
\end{cases}
\end{equation}

We used a total of 3 cycles over 1200 epochs, with $T_c = 400$ epochs per cycle. The model was trained on a workstation equipped with an NVIDIA GeForce RTX 3090 GPU (24 GB VRAM) with a batch size of 15, $a_r = 0.1$, $a_0 = 0.01$, $\gamma = 0.8$, and $\epsilon = 0.9$. The training required approximately 13 hours. 

\subsection{Evaluation}

To comprehensively evaluate the performance of the model, we assessed both segmentation accuracy and uncertainty calibration.

\textbf{Segmentation Metrics.}  
We report four standard metrics commonly used in medical image segmentation: Mean Dice Similarity Coefficient (mDice), Mean Intersection over Union (mIoU), Mean 95th percentile Hausdorff Distance (mHD95), and Mean Pixel Accuracy (mPA).

These metrics are used to compare our method with a range of baseline models, including the standard nnUNet-v2 \cite{isensee2021nnu}, UNet \cite{ronneberger2015u}, Attention UNet \cite{oktay2018attention}, TransUNet \cite{chen2021transunet}, DenseASPP \cite{yang2018denseaspp}, and FCN with ResNet101 backbone \cite{dai2016r}; nnUNet-v2 was trained using its default protocol with 5-fold cross-validation and ensembling. The metrics of other models were sourced from Wang et al. \cite{wang2024prediction}. 

\textbf{Uncertainty Calibration.}  
To assess the reliability of the model’s uncertainty estimates, we computed the Uncertainty Calibration Error (UCE) and visualized reliability diagrams. UCE evaluates how well predicted uncertainty aligns with observed prediction errors. It was computed by binning the uncertainty values and comparing the average uncertainty to the empirical error rate within each bin.

Formally, for each bin \( b \), we compute the average predicted uncertainty \( \bar{u}_b \) and empirical error rate \( \bar{e}_b \), and define UCE as:

\begin{equation}
\text{UCE} = \sum_{b=1}^{B} \frac{|S_b|}{\sum_{j=1}^{B} |S_j|} \cdot \left| \bar{u}_b - \bar{e}_b \right|,
\end{equation}

where \( S_b \) is the set of pixels in bin \( b \), and \( B \) is the total number of bins. We apply this evaluation using both entropy and standard deviation as the uncertainty scoring functions.

\section{Experimental Results}

The segmentation performance of seven models was evaluated on the test set described in Section~\ref{subsec:dataset} using four metrics: Mean Dice Similarity Coefficient (mDice), Mean Intersection over Union (mIoU), Mean 95th percentile Hausdorff Distance (mHD95), and Mean Pixel Accuracy (mPA). As shown in Table~\ref{tab:segmentation-results}, the standard nnUNet achieved the highest mDice (0.816) and mIoU (0.722), along with strong performance in mHD95 (94) and mPA (0.868). TransUNet performed best in mHD95 (89) and mPA (0.880), while our proposed Bayesian variant, nnUNet-B, achieved competitive results with an mDice of 0.805, mIoU of 0.709, mHD95 of 97, and mPA of 0.860. Attention UNet, DenseASPP, and FCN performed reasonably but fall short of the nnUNet variants in overlap and boundary accuracy. Overall, nnUNet-B demonstrated competitive segmentation performance while offering the added benefit of uncertainty estimation.

\begin{table}[h]
\setlength{\tabcolsep}{12pt}
\centering
\caption{Comparison of segmentation performance across various models. Metrics include Mean Dice Similarity Coefficient (\textbf{mDice}), Mean Intersection over Union (\textbf{mIoU}), Mean 95th percentile Hausdorff Distance (\textbf{mHD95}), and Mean Pixel Accuracy (\textbf{mPA}). The best results are highlighted in \textbf{bold}. \cite{wang2024prediction}}
\label{tab:segmentation-results}
\begin{tabular}{lcccc}
\toprule
\textbf{Model} & \textbf{mDice} & \textbf{mIoU} & \textbf{mHD95} & \textbf{mPA} \\
\midrule
\underline{nnUNet-B}            & 0.805  & 0.709  & 97  & 0.860 \\
nnUNet-v2                       & \textbf{0.816}  & \textbf{0.722}  & 94  & 0.868 \\
TransUNet                       & 0.800  & 0.720  & \textbf{89}  & \textbf{0.880} \\
UNet                            & 0.773  & 0.684  & 101 & 0.863 \\
Attention UNet                  & 0.787  & 0.696  & 101 & 0.787 \\
DenseASPP                       & 0.773  & 0.686  & 100 & 0.866 \\
FCN (ResNet101)                 & 0.783  & 0.692  & 99  & 0.868 \\
\bottomrule
\end{tabular}
\end{table}

\begin{figure}[ht!]
    \centering
    \includegraphics[width=\textwidth]{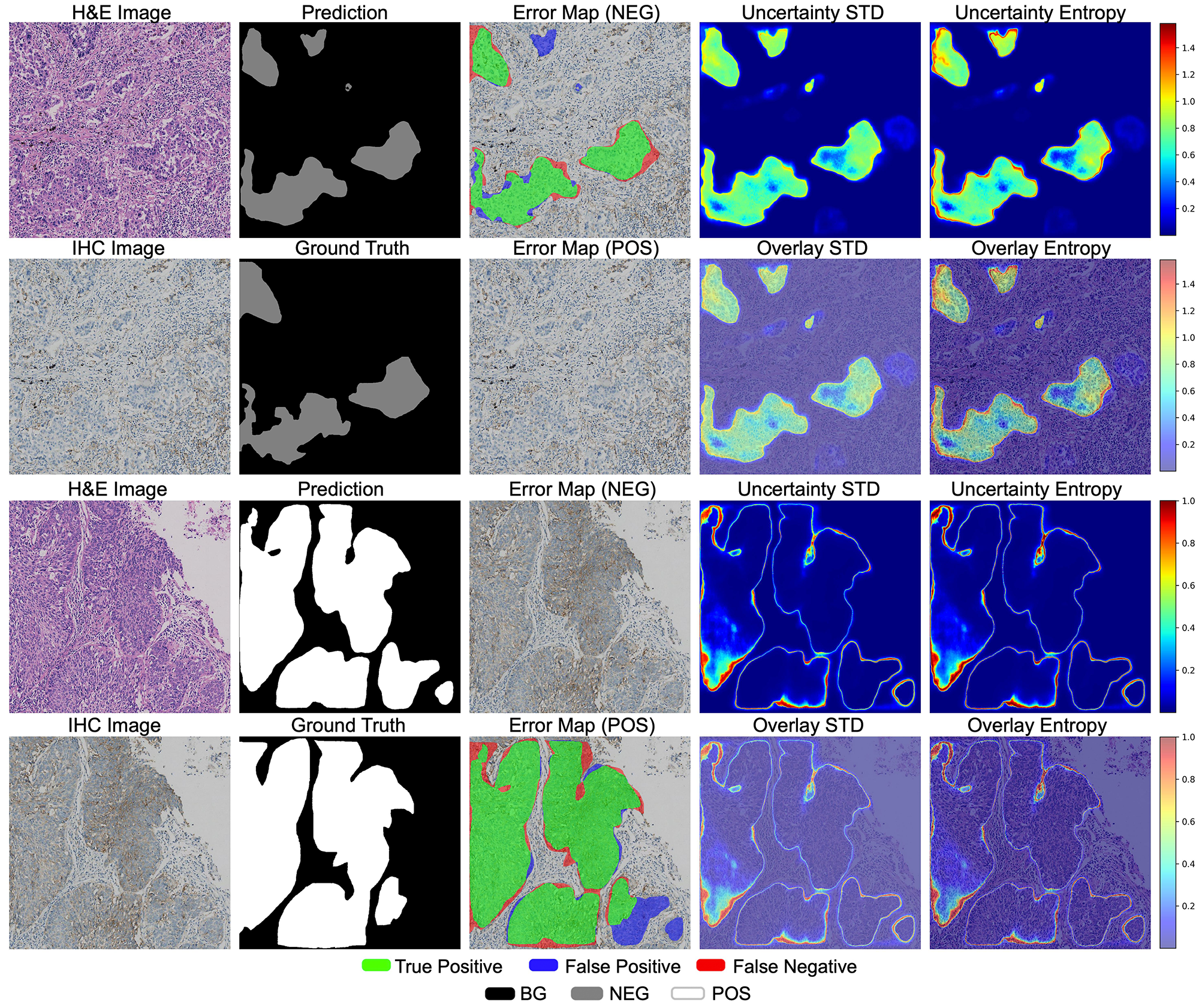}
    \caption{Visual summary of nnUNet-B predictions, error maps, and uncertainty estimates for two test images (top two rows: image 1; bottom two rows: image 2). For each image: Column 1 shows the H\&E and corresponding IHC reference; Column 2 displays the model prediction and ground truth; Column 3 presents class-specific error maps for PD-L1-negative (NEG) and -positive (POS) regions; Columns 4 and 5 show standard deviation and entropy-based uncertainty maps, each overlaid on the H\&E image.}    
    \label{fig:img_results}
\end{figure}

Fig.~\ref{fig:img_results} shows two examples of segmentation predictions from the test set, along with errors and uncertainty estimates. The predicted masks match the ground truth in both PD-L1-negative and positive regions, including complex cellular architecture. The error maps show few false positives or negatives, indicating good agreement with expert annotations. The model also successfully segments regions with mixed or ambiguous morphology, where different cell types are densely interwoven. The uncertainty maps show elevated uncertainty along region boundaries and in areas with heterogeneous or atypical cell morphology. 

\begin{figure}[ht]
    \centering
    \includegraphics[width=\textwidth]{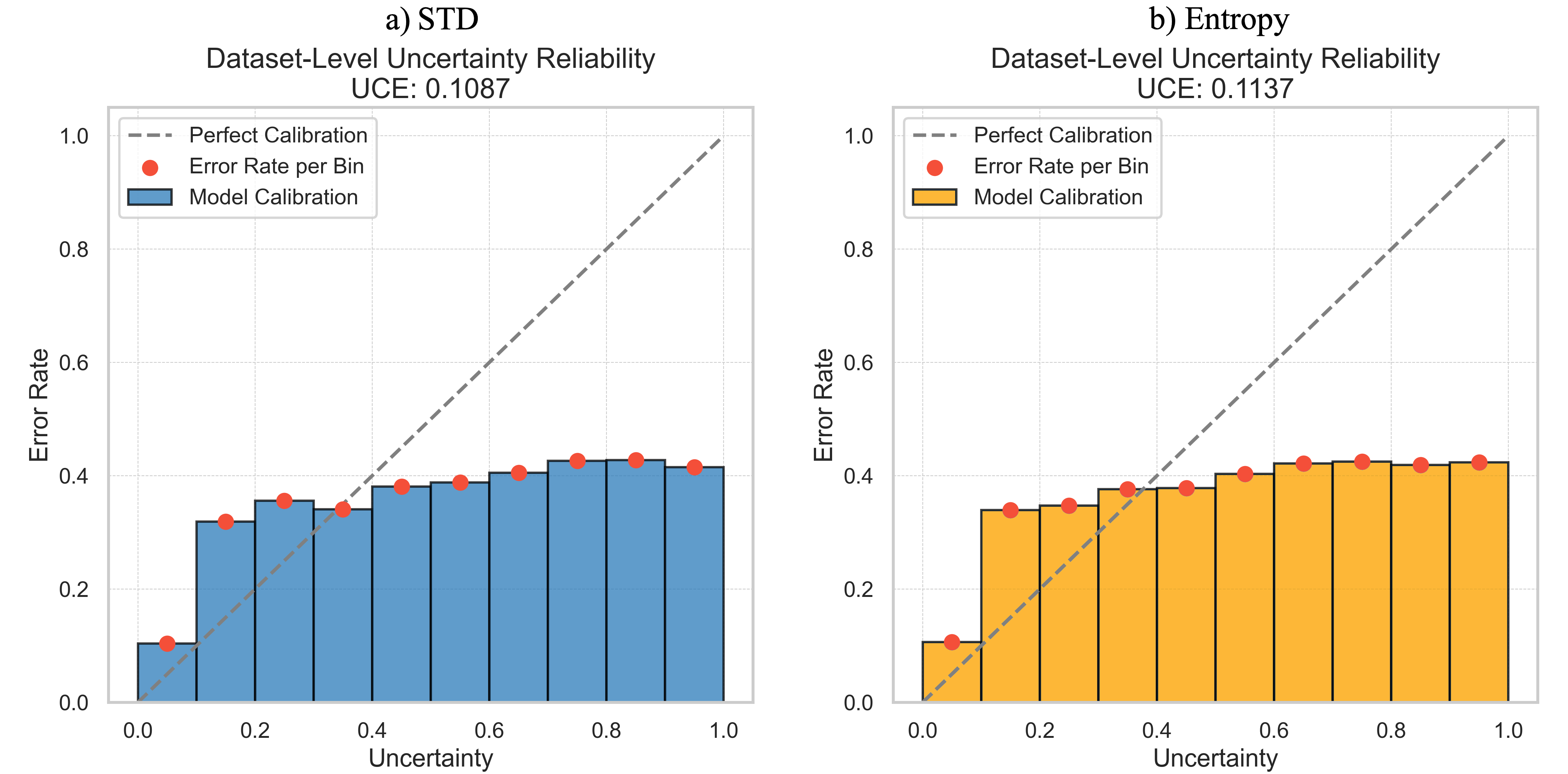}
    \caption{Test dataset-level uncertainty calibration curves using (a) STD and (b) entropy.
Each plot displays the relationship between predicted uncertainty and actual prediction error, computed across binned uncertainty intervals. The dashed diagonal line denotes perfect calibration, where predicted uncertainty would match the observed error.}
    \label{fig:uncertainty}
\end{figure}

Both STD and entropy demonstrate a clear positive correlation between predicted uncertainty and segmentation error, as shown in Fig.~\ref{fig:uncertainty}. However, both measures display miscalibration (most notably in the mid-to-high uncertainty bins) where the predicted uncertainty exceeds the actual error. The mean UCE is slightly lower for STD (0.1087) than for entropy (0.1137).

\section{Discussion}
This study demonstrates that PD-L1 expression can be inferred directly from H\&E-stained images using an uncertainty-aware segmentation framework based on nnUNet-v2~\cite{isensee2021nnu} and MPS~\cite{zhao2022efficient}. By sampling checkpoints during cyclic training, we approximate the model posterior without architectural changes or stochastic inference, offering a practical and interpretable alternative for histopathology tasks.

The proposed model (nnUNet-B) achieved strong performance across all segmentation metrics, with an mDice of 0.805, mIoU of 0.709, mHD95 of 97, and mPA of 0.860. While the regular nnUNet-v2 \cite{isensee2021nnu} slightly outperforms it in mDice (0.816) and mIoU (0.722), and achieved lower mHD95 (94); nnUNet-B provides comparable accuracy with the added benefit of reliable uncertainty quantification. These results confirm that incorporating MPS does not substantially compromise segmentation performance while enriching model outputs with interpretable confidence estimates.

Uncertainty measures based on standard deviation and entropy were well correlated with segmentation error (Fig.~\ref{fig:uncertainty}), but showed underestimated error in mid-to-high bins, which suggests an overestimation of risk in these regions. The UCE was slightly lower for STD (0.1087) than for entropy (0.1137), suggesting that STD better aligns with observed error and may be preferable in downstream applications.

The cyclic learning rate with polynomial decay was essential for encouraging checkpoint diversity while preserving convergence. Sampling checkpoints during the low-learning-rate phase of each cycle proved effective for posterior approximation and stable inference. Another advantage of this method is its versatility: in contrast to MCDO~\cite{gal2016dropout}, MPS can be applied to any segmentation model without the need to modify the backbone network's architecture.

Despite promising results, several limitations remain. The model was evaluated on a single cancer subtype (lung squamous cell carcinoma) using IHC-derived annotations, and its generalizability to other tissue types or biomarkers remains to be tested. Moreover, while uncertainty maps enhance interpretability, real-world utility will depend on integration with clinical workflows and further human-in-the-loop validation.

Future work should explore modality expansion during training, improved calibration techniques, domain adaptation across cancer types, and interactive decision-support systems that incorporate uncertainty estimates.

\section{Conclusions}
We propose a Bayesian segmentation framework using Multimodal Posterior Sampling (MPS) to infer PD-L1 expression from H\&E-stained images. By exploiting cyclic training and sampling diverse checkpoints, our model provides accurate segmentation with pixel-wise epistemic uncertainty estimates via entropy and standard deviation. It performs competitively with state-of-the-art methods while offering improved interpretability. This supports the feasibility of H\&E-based PD-L1 inference as a scalable alternative to IHC. Future directions include enhancing calibration, improving generalization, and integrating with clinical decision-making tools.

\section*{Acknowledgements}
This work was partially supported under grant PID2023-152631OB-I00 by the Ministerio de Ciencia, Innovación y Universidades, Agencia Estatal de Investigación  (MCIN/AEI/10.13039/501100011033/), \\
co-financed by European Regional Development Fund (ERDF), ’A way of making Europe’. Roman Kinakh holds a UC3M fellowship PIPF Programa "Inteligencia Artificial" \\
(Call 2024/2025).
Some icons used in figures were sourced from Flaticon.com and are attributed to their respective authors.
%
%
%
\bibliographystyle{elsarticle-num}
\bibliography{references.bib}

@article{topalian2012safety,
  title={Safety, activity, and immune correlates of anti--PD-1 antibody in cancer},
  author={Topalian, Suzanne L and Hodi, F Stephen and Brahmer, Julie R and Gettinger, Scott N and Smith, David C and McDermott, David F and Powderly, John D and Carvajal, Richard D and Sosman, Jeffrey A and Atkins, Michael B and others},
  journal={New England Journal of Medicine},
  volume={366},
  number={26},
  pages={2443--2454},
  year={2012},
  publisher={Mass Medical Soc}
}

@article{wang2024prediction,
  title={Prediction of PD-L1 tumor positive score in lung squamous cell carcinoma with H\&E staining images and deep learning},
  author={Wang, Qiushi and Deng, Xixiang and Huang, Pan and Ma, Qiang and Zhao, Lianhua and Feng, Yangyang and Wang, Yiying and Zhao, Yuan and Chen, Yan and Zhong, Peng and others},
  journal={Frontiers in Artificial Intelligence},
  volume={7},
  pages={1452563},
  year={2024},
  publisher={Frontiers Media SA}
}

@inproceedings{zhao2022efficient,
  title={Efficient Bayesian uncertainty estimation for nnU-Net},
  author={Zhao, Yidong and Yang, Changchun and Schweidtmann, Artur and Tao, Qian},
  booktitle={International Conference on Medical Image Computing and Computer-Assisted Intervention},
  pages={535--544},
  year={2022},
  organization={Springer}
}

@article{deng2025mcranet,
  title={MCRANet: MTSL-based connectivity region attention network for PD-L1 status segmentation in H\&E stained images},
  author={Deng, Xixiang and Luo, Jiayang and Huang, Pan and He, Peng and Li, Jiahao and Liu, Yanan and Xiao, Hualiang and Feng, Peng},
  journal={Computers in Biology and Medicine},
  volume={184},
  pages={109357},
  year={2025},
  publisher={Elsevier}
}

@article{isensee2021nnu,
  title={nnU-Net: a self-configuring method for deep learning-based biomedical image segmentation},
  author={Isensee, Fabian and Jaeger, Paul F and Kohl, Simon AA and Petersen, Jens and Maier-Hein, Klaus H},
  journal={Nature methods},
  volume={18},
  number={2},
  pages={203--211},
  year={2021},
  publisher={Nature Publishing Group}
}

@inproceedings{gal2016dropout,
  title={Dropout as a bayesian approximation: Representing model uncertainty in deep learning},
  author={Gal, Yarin and Ghahramani, Zoubin},
  booktitle={international conference on machine learning},
  pages={1050--1059},
  year={2016},
  organization={PMLR}
}

@inproceedings{smith2017cyclical,
  title={Cyclical learning rates for training neural networks},
  author={Smith, Leslie N},
  booktitle={2017 IEEE winter conference on applications of computer vision (WACV)},
  pages={464--472},
  year={2017},
  organization={IEEE}
}

@article{pardoll2012blockade,
  title={The blockade of immune checkpoints in cancer immunotherapy},
  author={Pardoll, Drew M},
  journal={Nature reviews cancer},
  volume={12},
  number={4},
  pages={252--264},
  year={2012},
  publisher={Nature Publishing Group UK London}
}

@article{fuchs2011computational,
  title={Computational pathology: challenges and promises for tissue analysis},
  author={Fuchs, Thomas J and Buhmann, Joachim M},
  journal={Computerized Medical Imaging and Graphics},
  volume={35},
  number={7-8},
  pages={515--530},
  year={2011},
  publisher={Elsevier}
}

@article{marini2021deep,
  title={Deep learning in histopathology: the path to the clinic},
  author={Van der Laak, Jeroen and Litjens, Geert and Ciompi, Francesco},
  journal={Nature medicine},
  volume={27},
  number={5},
  pages={775--784},
  year={2021},
  publisher={Nature Publishing Group US New York}
}

@article{tellez2021histo,
  title={HistoQC: an open-source quality control tool for digital pathology slides},
  author={Janowczyk, Andrew and Zuo, Ren and Gilmore, Hannah and Feldman, Michael and Madabhushi, Anant},
  journal={JCO clinical cancer informatics},
  volume={3},
  pages={1--7},
  year={2019},
  publisher={American Society of Clinical Oncology}
}

@article{stacke2021measuring,
  title={Measuring domain shift for deep learning in histopathology},
  author={Stacke, Karin and Eilertsen, Gabriel and Unger, Jonas and Lundstr{\"o}m, Claes},
  journal={IEEE journal of biomedical and health informatics},
  volume={25},
  number={2},
  pages={325--336},
  year={2020},
  publisher={IEEE}
}

@article{ribas2015releasing,
  title={Releasing the brakes on cancer immunotherapy},
  author={Ribas, Antoni and others},
  journal={N Engl J Med},
  volume={373},
  number={16},
  pages={1490--1492},
  year={2015}
}

@article{ai2020roles,
  title={Roles of PD-1/PD-L1 pathway: signaling, cancer, and beyond},
  author={Ai, Luoyan and Xu, Antao and Xu, Jie},
  journal={Regulation of cancer Immune checkpoints: Molecular and cellular mechanisms and therapy},
  pages={33--59},
  year={2020},
  publisher={Springer}
}

@inproceedings{ronneberger2015u,
  title={U-net: Convolutional networks for biomedical image segmentation},
  author={Ronneberger, Olaf and Fischer, Philipp and Brox, Thomas},
  booktitle={Medical image computing and computer-assisted intervention--MICCAI 2015: 18th international conference, Munich, Germany, October 5-9, 2015, proceedings, part III 18},
  pages={234--241},
  year={2015},
  organization={Springer}
}

@article{oktay2018attention,
  title={Attention u-net: Learning where to look for the pancreas},
  author={Oktay, Ozan and Schlemper, Jo and Folgoc, Loic Le and Lee, Matthew and Heinrich, Mattias and Misawa, Kazunari and Mori, Kensaku and McDonagh, Steven and Hammerla, Nils Y and Kainz, Bernhard and others},
  journal={arXiv preprint arXiv:1804.03999},
  year={2018}
}

@article{chen2021transunet,
  title={Transunet: Transformers make strong encoders for medical image segmentation},
  author={Chen, Jieneng and Lu, Yongyi and Yu, Qihang and Luo, Xiangde and Adeli, Ehsan and Wang, Yan and Lu, Le and Yuille, Alan L and Zhou, Yuyin},
  journal={arXiv preprint arXiv:2102.04306},
  year={2021}
}

@inproceedings{yang2018denseaspp,
  title={Denseaspp for semantic segmentation in street scenes},
  author={Yang, Maoke and Yu, Kun and Zhang, Chi and Li, Zhiwei and Yang, Kuiyuan},
  booktitle={Proceedings of the IEEE conference on computer vision and pattern recognition},
  pages={3684--3692},
  year={2018}
}

@article{dai2016r,
  title={R-fcn: Object detection via region-based fully convolutional networks},
  author={Dai, Jifeng and Li, Yi and He, Kaiming and Sun, Jian},
  journal={Advances in neural information processing systems},
  volume={29},
  year={2016}
}

@inproceedings{isensee2024nnu,
  title={nnu-net revisited: A call for rigorous validation in 3d medical image segmentation},
  author={Isensee, Fabian and Wald, Tassilo and Ulrich, Constantin and Baumgartner, Michael and Roy, Saikat and Maier-Hein, Klaus and Jaeger, Paul F},
  booktitle={International Conference on Medical Image Computing and Computer-Assisted Intervention},
  pages={488--498},
  year={2024},
  organization={Springer}
}
\end{document}